\documentclass[runningheads]{llncs}
\usepackage[T1]{fontenc}
\usepackage{graphicx}
\usepackage{subfig}
\usepackage{xcolor}
\usepackage{amsmath}
\usepackage{dirtytalk}
\usepackage{booktabs}
\usepackage{amsfonts} 
\usepackage{multirow}
\usepackage{caption}
\usepackage{hyperref}
\usepackage[misc]{ifsym}
\hypersetup
{
colorlinks=true,
linkcolor=blue,
filecolor=blue,
urlcolor=blue,
}

\begin{document}
\title{LUCYD: A Feature-Driven Richardson-Lucy Deconvolution Network}
\titlerunning{LUCYD}
%

\author{Tomáš Chobola\inst{1,4}\orcidID{0009-0000-3272-9996} \and Gesine Müller\inst{2} \and Veit Dausmann\inst{3}\orcidID{0000-0003-3281-9208} \and Anton Theileis\inst{3} \and Jan Taucher\inst{3}\orcidID{0000-0001-9944-0775} \and Jan Huisken\inst{2}\orcidID{0000-0001-7250-3756} \and Tingying Peng\inst{4}\orcidID{0000-0002-7881-1749}\textsuperscript{\Letter}}
\index{Chobola, Tomáš}
\index{Müller, Gesine}
\index{Dausmann, Veit}
\index{Theileis, Anton}
\index{Taucher, Jan}
\index{Huisken, Jan}
\index{Peng, Tingying}

\authorrunning{T. Chobola et al.}

\institute{Technical University of Munich, Munich, Germany \and Georg-August-University Göttingen, Göttingen, Germany \and GEOMAR Helmholtz Centre for Ocean Research Kiel, Kiel, Germany \and Helmholtz AI, Helmholtz Munich - German Research Center for Environmental Health, Neuherberg, Germany\\
\email{tingying.peng@helmholtz-muenchen.de}}

%
%
\maketitle              
\begin{abstract}
The process of acquiring microscopic images in life sciences often results in image degradation and corruption, characterised by the presence of noise and blur, which poses significant challenges in accurately analysing and interpreting the obtained data. This paper proposes LUCYD, a novel method for the restoration of volumetric microscopy images that combines the Richardson-Lucy deconvolution formula and the fusion of deep features obtained by a fully convolutional network. By integrating the image formation process into a feature-driven restoration model, the proposed approach aims to enhance the quality of the restored images whilst reducing computational costs and maintaining a high degree of interpretability. Our results demonstrate that LUCYD outperforms the state-of-the-art methods in both synthetic and real microscopy images, achieving superior performance in terms of image quality and generalisability. We show that the model can handle various microscopy modalities and different imaging conditions by evaluating it on two different microscopy datasets, including volumetric widefield and light-sheet microscopy. Our experiments indicate that LUCYD  can significantly improve resolution, contrast, and overall quality of microscopy images. Therefore, it can be a valuable tool for microscopy image restoration and can facilitate further research in various microscopy applications. We made the source code for the model accessible under \href{https://github.com/ctom2/lucyd-deconvolution/}{https://github.com/ctom2/lucyd-deconvolution/}.

\end{abstract}

\section{Introduction}

Microscopy is one of the most widely used imaging techniques that allows life scientists to analyse cells, tissues and subcellular structures with a high level of detail. However, microscopy images often suffer from degradation such as blur, noise and other artefacts, which may result in an inaccurate quantification and hinder downstream analysis. Therefore, deconvolution techniques are necessary to restore the images to improve their quality, thus increasing the accuracy of downstream tasks \cite{Wallace2001,Sage2017,9268933}. Image deconvolution is a well-studied task in computer vision and imaging sciences that aims to recover a sharp and clear object out of a degraded input. The mathematical representation of image corruption can be expressed as:
\begin{equation}
    y=x*K+n,
\end{equation}
where $*$ represents convolution, $y$ denotes the resulting image of an object $x$, which has been blurred with a point spread function (PSF) $K$, and degraded by noise $n$.

Two classic image deconvolution methods widely used in microscopy and medical imaging are Wiener filter \cite{wiener1949extrapolation} and Richardson-Lucy algorithm (RL) \cite{Richardson:72,Lucy1974}. The Wiener filter is a linear filter that is applied to the frequency domain representation of the blurred image. It assumes the Gaussian noise distribution and thus minimises the mean squared error between the restored image and the original one. The RL method, on the other hand, is an iterative algorithm that works in the spatial domain, usually leading to better reconstruction than Wiener filter. 
It assumes Poisson noise distribution and seeks to estimate the corresponding sharp image $x$ in a fixed number of iterations or until a convergence criterion is met \cite{eichstadt2013comparison}. While being simple and effective, the main limitations of both methods are their susceptibility to noise amplification \cite{Tan2018,DELLACQUA20101446} and the assumption that the accurate PSF is known. In practice, however, PSF is challenging to obtain and is often unknown or varies across the image, which leads to inaccurate reconstructions of the sharp image. Moreover, as an iterative method, RL is computationally costly for three-dimensional (3D) data \cite{Dey2006}.

In the computer vision field, numerous deep learning models have been trained on large datasets with the objective to learn a direct mapping between input and output domains \cite{Guo2020,Chen2021,Qiao2021,Vizcaino2021,Wagner2021}. Some of these models have also been adapted for use in microscopy, such as the U-Net-based content-aware image restoration networks (CARE) \cite{Weigert2018}. These methods have exhibited exceptional performance in tasks such as super-resolution and denoising. However, the interpretability of these methods is limited, and given their data-driven nature, the quantity and quality of training data can be a restricting factor, particularly in biomedical applications where data pairs are often scarce or not available. 

Inspired by the RL algorithm, the Richardson-Lucy Network (RLN) \cite{Li2022} was recently designed to overcome the problem of data-driven models by embedding the RL formula for iterative image restoration into a neural network and substituting convolutions of the measured PSF kernel with learnable convolutional layers. Although being more compact than U-Net, the low capacity of RLN makes it insufficiently robust to different blur intensities and noise levels, requiring the network to be re-trained whenever there is a shift in the input image domain. This reduces the efficacy of the method.

To address the limitations of existing methods, we propose a novel lightweight model called LUCYD, which integrates the RL deconvolution formula and a U-shaped network. The main contributions of this paper can be summarised as:
\begin{enumerate}
    \item LUCYD is a lightweight deconvolution method that embeds the RL deconvolution formula into a deep convolutional network that leverages the features extracted by a U-shaped module while maintaining low computational costs for processing 3D microscopy images and a high level of interpretability. 
    \item The proposed method outperforms existing deconvolution methods on both real and synthetic datasets, based on qualitative assessment and quantitative evaluation metrics, respectively.
    \item We show that LUCYD has strong resistance to noise and can generalise well to new and unseen data. This ability makes it a valuable tool for practical applications in microscopy imaging fields where image quality is critical for downstream tasks yet training data are often scarce or unavailable.
\end{enumerate}

\section{Method}

\begin{figure}[t]
    \centering
    \includegraphics[width=.9\textwidth]{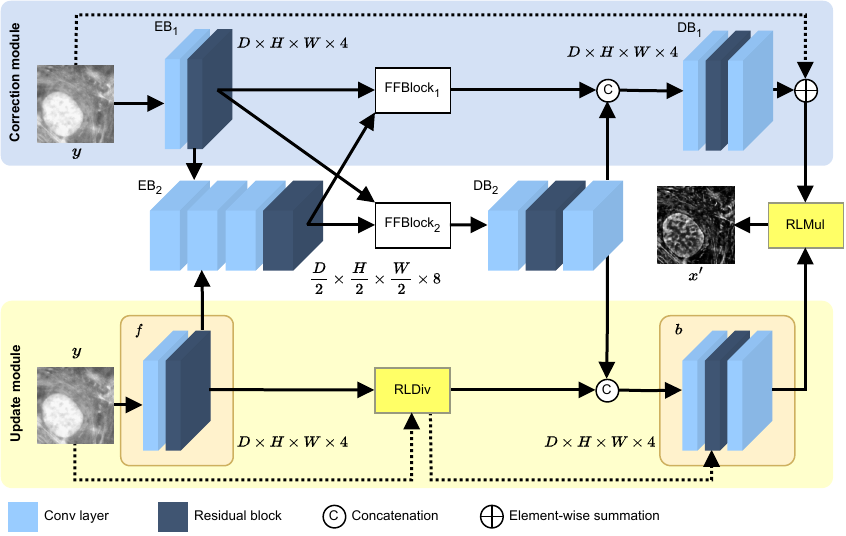}
    \caption{The architecture of LUCYD consists of a correction module, an update module and a bottleneck that is shared between the two modules.}
    \label{fig:architecture}
\end{figure}

The overall architecture of the proposed model is illustrated in Figure \ref{fig:architecture} and it comprises three main components: a \textbf{correction module}, an \textbf{update module}, and a \textbf{bottleneck} that is shared between the two modules. The data flow in the model is based on the following iterative RL formula:
\begin{equation}\label{eq:rl}
    z_{(k)}=\underbrace{\vphantom{\left(\frac{0}{0_{(x)}}\right)}z_{(k-1)}}_{x\text{ estimate}}\cdot\underbrace{\left(\frac{y}{z_{(k-1)}*K}*K^\top\right)}_{\text{update term}},
\end{equation}
which aims to recover $x$ in $k$ steps. We bypass the requirement of $k{-}1$ preceding iterations with the correction module that generates a mask $M$ to form an intermediate sharp image estimation through a single forward pass, allowing to rapidly process 3D data, as follows:
\begin{equation}
    \tilde{z}=y+M.
\end{equation}

Next, inspired by Li \textit{et al.} \cite{Li2022}, we adopt the three-step approach to decompose the RL update term from Equation \ref{eq:rl} in the update module:
\begin{equation}
    \text{(a)}\;\text{FP}=y*f\text{, (b)}\;\text{DV}={y}/\text{FP}\text{, (c)}\;u=\text{DV}*b.
\end{equation}
Specifically, we replace convolutions with a known PSF in steps (a) and (c) with forward projector $f$ and backward projector $b$, which consist of sets of learnable convolutional layers. The produced update term $u$ allows us to recondition the estimate $\tilde{z}$ from the correction module into a sharp image through multiplication, i.e. the last step of image formation in the RL formula: $x'=\tilde{z}\cdot u$. The whole network can then be expressed as follows,
\begin{equation}\label{eq:lucyd}
    x'=\tilde{z}\cdot\left(\frac{y}{y*f}*b\right).
\end{equation}
By adhering to the image formation steps as prescribed by the RL formula, we maintain a high degree of interpretability, critical for real-world scenarios, where the accuracy and reliability of the generated results are of utmost importance.

\begin{figure}[h]
    \centering
    \subfloat[]{
        \label{fig:ffblock}
        \centering
        \includegraphics[width=.4\linewidth]{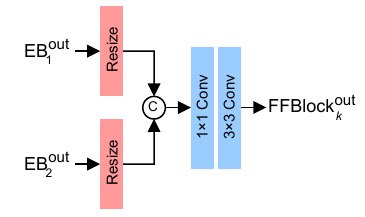}
    }
    \subfloat[]{
        \label{fig:rldiv}
        \centering
        \includegraphics[width=.4\linewidth]{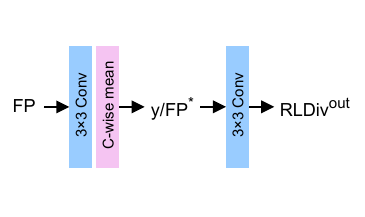}
    }
\caption{The architecture of submodules: (a) Feature Fusion Block (FFBlock), (b) Richardson-Lucy Division Block (RLDiv).}
\label{fig:lucyd_blocks}
\end{figure}

\subsection{Correction Module \& Bottleneck}

The proposed correction module and bottleneck architectures consist of encoder blocks (EBs), decoder blocks (DBs), and multi-scale feature fusion blocks to facilitate efficient information exchange across different scales within the model.

\subsubsection{Feature Encoding} 

The features of the volumetric input image $y\in\mathbb{R}^{C\times D\times H\times W}$ are obtained through the first encoder block $\text{EB}_1$ in the correction module, and then encoded by a convolutional layer with a stride 2. Subsequently, the downsampled features are concatenated with the encoded features of the forward projection $f$ from the update module and then fed to the bottleneck encoder $\text{EB}_2$ to integrate the information from both modules.

\subsubsection{Feature Fusion Block}

Similarly to Cho \textit{et al.} \cite{cho2021rethinking}, we enhance the connections between encoders and decoders and allow information flow from different scales within the network through Feature Fusion Blocks (FFBlocks). The features from $\text{EB}_1$ and $\text{EB}_2$ are refined as follows,
\begin{align}
    \text{FFBlock}_1^\text{out}&=\text{FFBlock}_1\left(\text{EB}_1^\text{out},(\text{EB}_2^\text{out})^\uparrow\right),\\
    \text{FFBlock}_2^\text{out}&=\text{FFBlock}_2\left((\text{EB}_1^\text{out})^\downarrow,\text{EB}_2^\text{out}\right),
\end{align}
where up-sampling ($\uparrow$) and down-sampling ($\downarrow$) is applied to allow for feature concatenation. The multi-scale features are then combined and processed by $1\times 1$ and $3\times 3$ convolutional layers, respectively, to allow the decoder blocks $\text{DB}_1$ and $\text{DB}_2$ to utilise information obtained on different scales. The structure of the blocks is shown in Figure \ref{fig:ffblock}.

\subsubsection{Feature Decoding}

Initially, the refined features are decoded in the bottleneck using a convolutional layer and residual block within the $\text{DB}_2$. Next, these features are expanded with a convolutional layer to match the dimensions in both the correction and update modules. The resulting features are then concatenated with the output of $\text{FFBlock}_1$ and subsequently fed into decoder $\text{DB}_2$ within the correction module. The features are then mapped to the image dimensions resulting in mask $M$, which is summed with $y$ to form $\tilde{z}$.

\subsection{Update Module}

Inspired by the forward and backward projector functions \cite{Li2022}, we substitute the PSF convolution operations from Richardson-Lucy algorithm with learnable convolutional layers and residual blocks. 

During forward projection (FP), shallow features are initially extracted by a single convolutional layer and then refined by a residual block. The output of $f$ is then passed to Richardson-Lucy Division Block (RLDiv) which embeds the division of the raw image $y$ by the channel-wise mean of the refined FP features. Next, we project the division result to a feature map to extract more information about the image. The visualisation of the process is in Figure \ref{fig:rldiv}. These features are then concatenated with the features extracted by the bottleneck and combined by a convolutional layer which initiates the backward projection with $b$. The output is then summed with the output of RLDiv, forming a skip-connection, and passed through a residual block. The features are then refined by a convolutional layer and their channel-wise mean is taken to be the \say{update term} $u$, which is used to obtain the final model output $x'$ through multiplication with $\tilde{z}$ (denoted as RLMul).

\subsection{Loss Function}

The entire model is trained end-to-end with a single loss function that combines the Mean Squared Error (MSE) and the Structural Similarity Index Measure (SSIM) as follows:
\begin{equation}
    \mathcal{L}(x',x)=\text{MSE}(x',x) - \ln\left(\frac{1 + \text{SSIM}(x',x)}{2}\right),
\end{equation}
where $x$ is the ground truth sharp image and $x'$ is the model estimation of $x$.

\section{Experiments}

\begin{figure}[t]
    \centering
    \subfloat[$y$]{
        \label{fig:interpetability_y}
        \centering
        \includegraphics[width=0.16\linewidth]{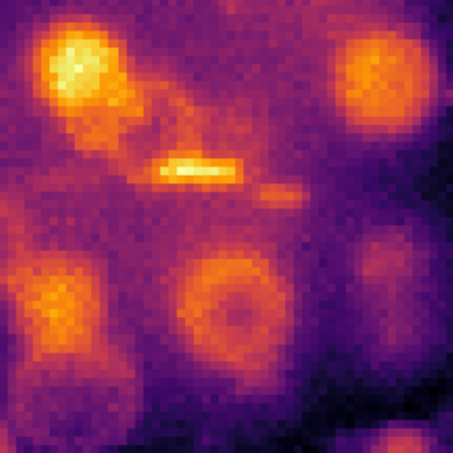}
    }
    \subfloat[$\tilde{z}$]{
        \label{fig:interpetability_z}
        \centering
        \includegraphics[width=0.16\linewidth]{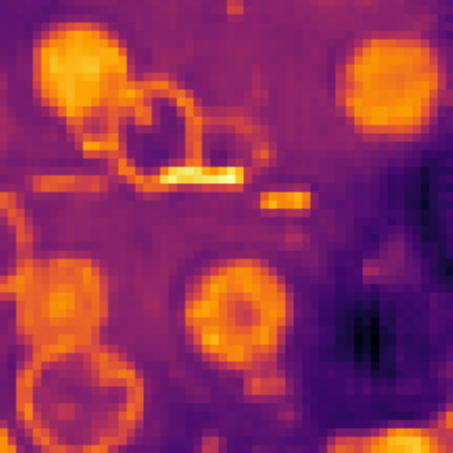}
    }
    \subfloat[$u$]{
        \label{fig:interpetability_update}
        \centering
        \includegraphics[width=0.16\linewidth]{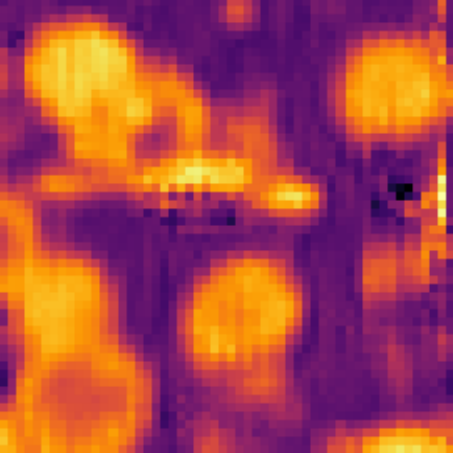}
    }
    \subfloat[$x'$]{
        \label{fig:interpetability_x_prime}
        \centering
        \includegraphics[width=0.16\linewidth]{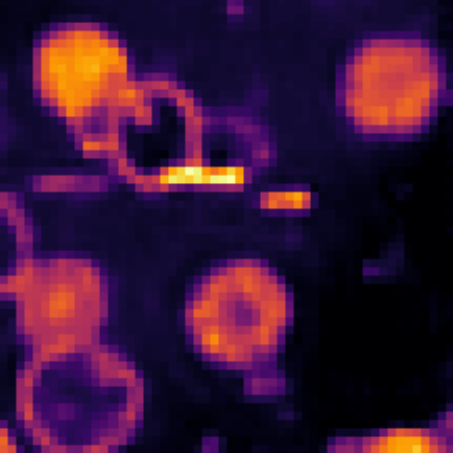}
    }
    \subfloat[$x$ (GT)]{
        \label{fig:interpetability_x}
        \centering
        \includegraphics[width=0.16\linewidth]{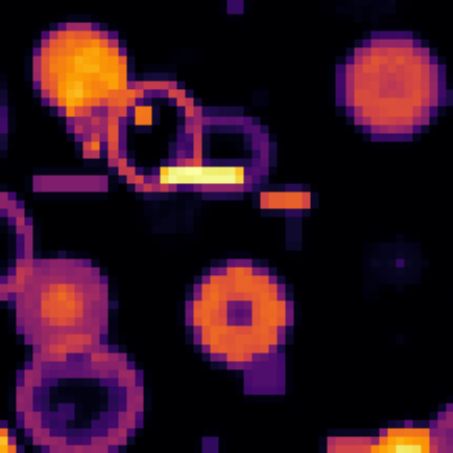}
    }
\caption{Overview of the deconvolution process given an input (a). The outputs of the correction module and the update module are shown in (b) and (c), respectively, and the final output obtained through their multiplication is in (d).}
\label{fig:interpretability}
\end{figure}

\subsection{Setup}

\begin{table}[b]
\centering
\caption{Number of learnable parameters, comparing CARE, RLN and LUCYD.}
\begin{tabular}{c|c|c}
\toprule
CARE \cite{Weigert2018} & RLN \cite{Li2022} & LUCYD (ours) \\
\hline\hline
1 M    & 15,900  & 24,964  \\
\bottomrule
\end{tabular}
\end{table}

\subsubsection{Datasets} We assess the performance of LUCYD on both simulated phantom objects and real microscopy images. To achieve this, we use five sets of 3D grayscale volumes generated by Li \textit{et al.} \cite{Li2022}, consisting of dots, solid spheres, and ellipsoidal surfaces, which are provided along with their sharp ground truth volumes of dimensions $128\times 128\times 128$ (one exemplary image shown in Figure \ref{fig:interpetability_x}). To test the generalization capabilities of our method, we also include two blurry and noisy versions of the dataset, $\mathcal{D}_\text{nuc}$ and $\mathcal{D}_\text{act}$, which utilize different image degradation processes for embryonic nuclei and membrane data. Additionally, we generate a mixed dataset by applying permutations of three Gaussian blur intensities ($\sigma_b=[1.0,1.2,1.5]$) and three levels of additive Gaussian noise ($\sigma_n=[0,15,30]$) to the ground truth volumes, and then test the ability of the model to generalize to volumes blurred with Gaussian kernels ($\sigma_b=[0.5,2.0]$) and additive Gaussian noise ($\sigma_n=[20,50,70,100]$) levels outside of the training dataset. The model is trained on patches of dimensions $32\times 64\times 64$ that are randomly sampled from the training datasets. Moreover, we evaluate the model trained using synthetic phantom shapes on a real 3D light-sheet image of a starfish (private data) and widefield microscopy image of U2OS cell (from the dataset of \cite{Li2022}), to explore the generalisation capabilities.

\subsubsection{Baseline \& Metrics} We employ one classic U-Net-based fluorescence image restoration model CARE \cite{Weigert2018} and one RL-based convolutional model RLN \cite{Li2022} as baselines. We quantitatively evaluate the deconvolution performance on simulated data using two metrics: Structural Similarity Index Measure (SSIM) and Peak Signal-to-Noise Ratio (PSNR).

\begin{table}[h]
\centering
\caption{Performance on synthetic datasets (SSIM/PSNR (dB)) degraded with blur and noise levels not present in the training dataset. The models are trained on phantom objects blurred with $\sigma_b=[1.0,1.2,1.5]$ and corrupted with Gaussian noise intensities $\sigma_n=[0,15,30]$.}
\label{tab:generlisation}
\begin{tabular}{c|c|c|c|c}
\toprule
Blur intensity $\sigma_b$ & Noise level $\sigma_n$ & CARE \cite{Weigert2018} & RLN \cite{Li2022} & LUCYD (ours)   \\ \hline\hline
0.5                     & 20                  & $0.9166/21.62$ & $0.9571/25.60$                     & $\mathbf{0.9725/26.85}$ \\
0.5                     & 50                  & $0.7589/15.96$ & $0.8519/21.67$                     & $\mathbf{0.9463/24.35}$ \\
0.5                     & 70                  & $0.6828/14.32$ & $0.7235/18.52$                     & $\mathbf{0.9040/21.78}$ \\
0.5                     & 100                 & $0.5856/12.56$ & $0.5644/15.91$                     & $\mathbf{0.7233/17.47}$ \\ \hline\hline
2.0                     & 20                  & $0.8582/20.36$ & $0.9040/22.34$                     & $\mathbf{0.9271/23.49}$ \\
2.0                     & 50                  & $0.7057/16.35$ & $0.7443/18.85$                     & $\mathbf{0.8575/21.00}$ \\
2.0                     & 70                  & $0.6259/15.06$ & $0.6051/16.69$                     & $\mathbf{0.7995/19.38}$ \\
2.0                     & 100                 & $0.5154/13.08$ & $0.4495/14.86$                     & $\mathbf{0.6311/16.42}$ \\ \bottomrule
\end{tabular}
\end{table}

\begin{table}[h]
\centering
\caption{Performance on synthetic datasets (SSIM/PSNR (dB)) given varying training data.}
\label{tab:nuc-act}
\begin{tabular}{l|c|c|ccc}
\toprule
                                 & Train dataset                         & Test dataset                         & \multicolumn{1}{c|}{CARE \cite{Weigert2018}}               & \multicolumn{1}{c|}{RLN \cite{Li2022}}                & \multicolumn{1}{c}{LUCYD (ours)} \\ \hline\hline
\multirow{2}{*}{In-domain}       & $\mathcal{D}_\text{nuc}$       & $\mathcal{D}_\text{nuc}$      & \multicolumn{1}{l|}{$0.7895/18.00$} & \multicolumn{1}{l|}{$0.9247/26.43$} & $\mathbf{0.9525/28.57}$ \\ 
                                 & $\mathcal{D}_\text{act}$       & $\mathcal{D}_\text{act}$      & \multicolumn{1}{l|}{$0.7666/17.44$} & \multicolumn{1}{l|}{$0.8966/26.10$} & $\mathbf{0.9450/27.83}$ \\ \hline\hline
\multirow{2}{*}{Cross-domain}    & $\mathcal{D}_\text{nuc}$       & $\mathcal{D}_\text{act}$      & \multicolumn{1}{l|}{$0.7623/17.68$} & \multicolumn{1}{l|}{$0.8841/24.33$} & $\mathbf{0.9024/24.82}$ \\ 
                                 & $\mathcal{D}_\text{act}$       & $\mathcal{D}_\text{nuc}$      & \multicolumn{1}{l|}{$0.7584/17.00$} & \multicolumn{1}{l|}{$0.9081/27.23$} & $\mathbf{0.9336/27.63}$ \\ \bottomrule
\end{tabular}
\end{table}

\begin{figure}[h!]
    \centering
    \includegraphics[width=.8\textwidth]{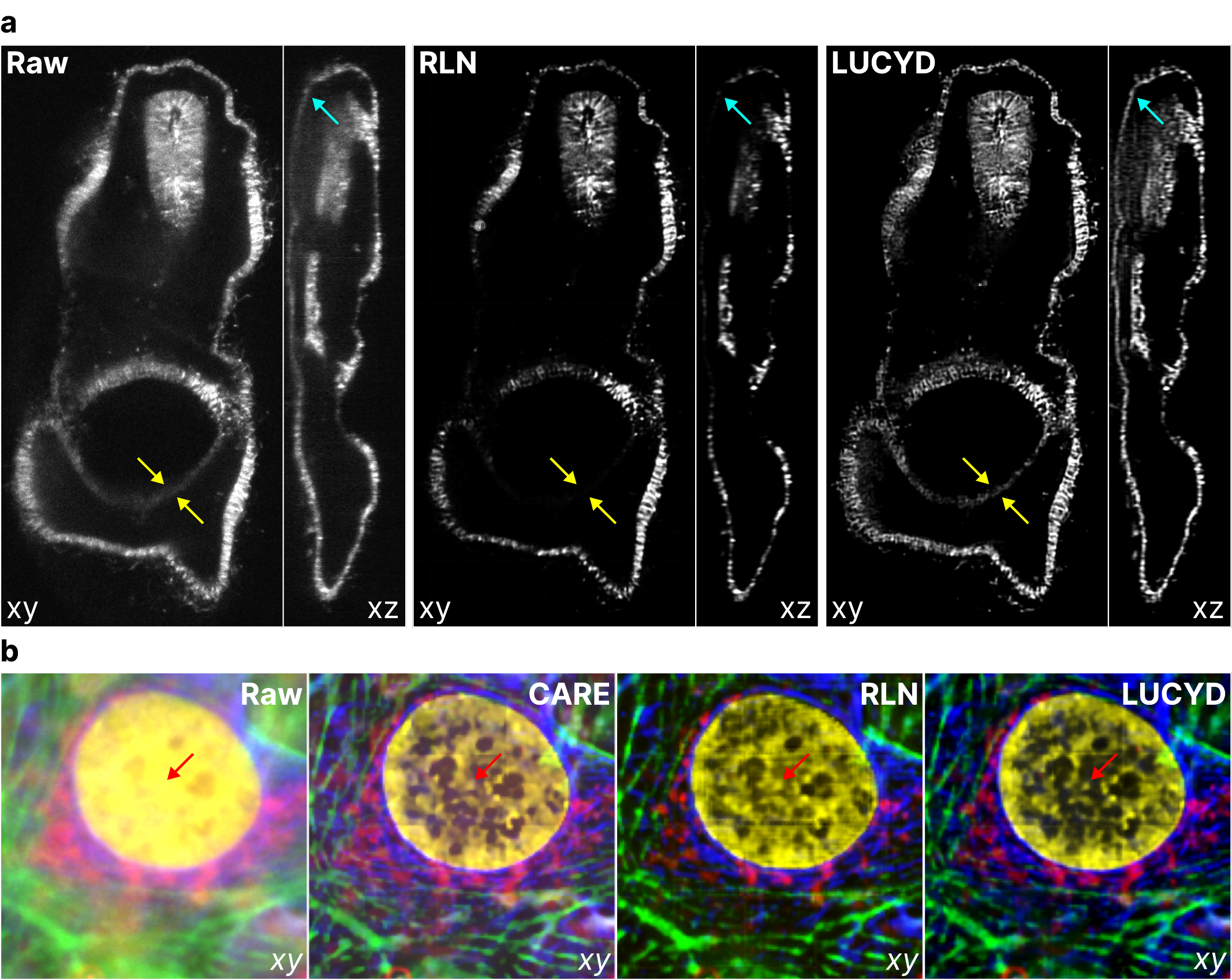}
    \caption{Quantitative comparison of RLN and LUCYD on lateral and axial maximum-intensity projections of a starfish acquired by 3D light-sheet microscopy is shown in (a). Additional analysis of the deconvolution results of CARE, RLN and LUCYD trained on synthetic phantom objects in (b) shows patches of four-colour lateral maximum-intensity projections of a fixed U2OS cell acquired by widefield microscopy (from the dataset of \cite{Li2022}). LUCYD exhibits superior performance in recovering fine details and structures as compared to CARE and RLN, while simultaneously maintaining low levels of noise and haze surrounding the objects.}
    \label{fig:qualitative}
\end{figure}

\subsection{Results}

In Table \ref{tab:generlisation}, we present the quantitative results of all three methods on simulated phantom objects degraded with blur and noise levels that were not present in the training dataset. LUCYD achieves the best performance even in cases where the amount of additive noise exceeds the maximum level included in the training dataset. This is in contrast to CARE and RLN, which did not demonstrate such exceptional generalisation capabilities and noise resistance. We further examine LUCYD's performance on datasets simulating widefield microscopy imaging of embryo nuclei and membrane data. As shown in Table \ref{tab:nuc-act}, LUCYD outperforms CARE and RLN in both in-domain and cross-domain assessments, further supporting the model's capabilities in cross-domain applications. 

We finally apply LUCYD on two real microscopy test samples, as illustrated in Figure \ref{fig:qualitative}. On the 3D light-sheet image of starfish, LUCYD recovers more details and structures than RLN while maintaining low levels of noise and haze surrounding the object in both lateral and axial projections. On the other test sample of a fixed U2OS cell acquired by widefield microscopy, LUCYD also suppresses noise and haze to a higher degree compared to RLN and CARE and retrieves finer and sharper details.


\section{Conclusion}

In this paper, we introduce LUCYD, an innovative technique for deconvolving volumetric microscopy images that combines a classic image deconvolution formula with a U-shaped network. LUCYD takes advantages of both approaches, resulting in a highly efficient method capable of processing 3D data with high efficacy. We have demonstrated through experiments on both synthetic and real microscopy datasets that LUCYD exhibits strong generalization capabilities, as well as robustness to noise. These qualities make it an excellent tool for cross-domain applications in various domains, such as biology and medical imaging. Additionally, the lightweight nature of LUCYD makes it computationally feasible for real-time applications, which can be crucial in various settings. 
%
%

\subsubsection{Acknowledgements} Tomáš Chobola is supported by the Helmholtz Association under the joint research school "Munich School for Data Science - MUDS".

%
%
%
%

\bibliographystyle{splncs04}
\bibliography{main}

\end{document}